\documentclass[10pt,twocolumn,letterpaper]{article}

\usepackage{cvpr}
\usepackage{times}
\usepackage{epsfig}
\usepackage{graphicx}
\usepackage{amsmath}
\usepackage{amssymb}
\usepackage{tabularx}
\usepackage[none]{hyphenat}

\cvprfinalcopy 

\def\cvprPaperID{1927} 

\begin{document}

\title{Object-Level Context Modeling For Scene Classification with Context-CNN}

\author{Syed Ashar Javed$^{1}$ and Anil Kumar Nelakanti$^{2}$\\
$^{1}$IIIT Hyderabad, $^{2}$Amazon\\
{\tt\small \{ashar.javed@research.iiit.ac.in,annelaka@amazon.com\}}
}

\maketitle


\begin{abstract}
    Convolutional Neural Networks (CNNs) have been used extensively for computer vision tasks and produce rich feature representation for objects or parts of an image. But reasoning about scenes requires integration between the low-level feature representations and the high-level semantic information. We propose a deep network architecture which models the semantic context of scenes by capturing object-level information. We use Long Short Term Memory(LSTM) units in conjunction with object proposals to incorporate object-object relationship and object-scene relationship in an end-to-end trainable manner. We evaluate our model on the LSUN dataset and achieve results comparable to the state-of-art. We further show visualization of the learned features and analyze the model with experiments to verify our model's ability to model context.
\end{abstract}

\section{Introduction} \label{introduction}

The task of classifying a scene requires assimilation of complex, inter-connected information. With the great success of large convolutional networks, deep features have replaced the low-level hand crafted features. CNN-only models for scene classification ~\cite{zhou2014learning,sharif2014cnn} show improvement in performance over methods using hand-engineered features and have been used to set baseline performance. But for challenging scenes, the holistic scene information is not distilled into the CNN model as the layers are locally connected and do not make use of the high order semantic context of the scene. Thus vanilla CNNs by design, are not suitable for capturing contextual knowledge like the complex interaction of objects in a scene. Other more sophisticated approaches from the recent literature either involve multiple networks with high number of parameters trained for weeks or models involving components which are learned separately. This either leads to models with very high complexity or models which incoherently fuse together information from different components, thus limiting the effectiveness of the training process.

\begin{figure}[t]
  \includegraphics[width=\linewidth]{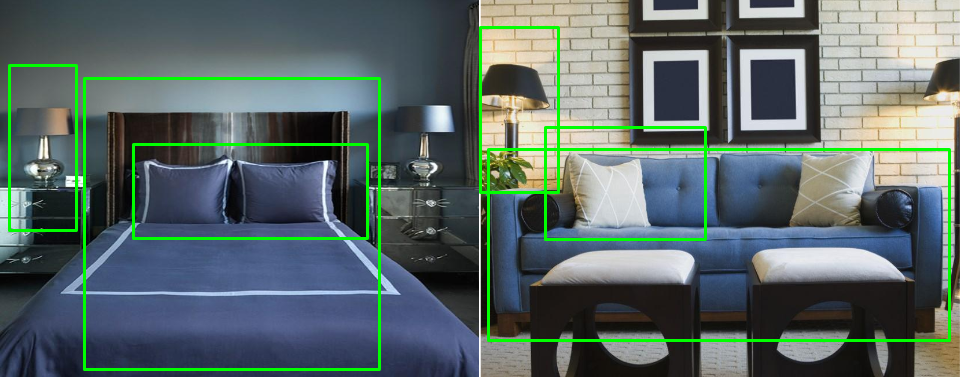}
  \caption{Distinguishing between complex scenes with similar global attributes and similar objects requires contextual reasoning. A bedroom scene and a living room scene both contain pillows and table lamps which by themselves are non-discriminative objects for their scene category even if their spatial position is taken into consideration.}
  \label{fig:fig1}
\end{figure}

In this work, we propose the Context-CNN model which encodes object-level context using object proposals and LSTM units on top of a CNN which extracts deep image features. This architecture attempts to bridge the semantic gap in scenes by modeling object-object and scene-object relationships within a single system. The intuition that the joint existence of a set of objects in a scene highly influences the final scene category has been regularly highlighted and verified empirically in various works in computer vision~\cite{li2010object,li2012object,liao2016understand}. Additionally, the LSTM units are capable of modeling the relationship between the objects by taking in object bounding boxes at each time step. In effect, this makes the network learn about the scene class probability distribution given it has seen a specific set of objects through time. For example, as shown in Figure~\ref{fig:fig1}, seeing a bed after a pillow cushion should hint towards the scene being a bedroom scene while seeing a sofa after a pillow cushion would suggest it's a living room scene. Note that even though the model captures object-level context, it does not need any labeled objects in the dataset as the objects are represented through their CNN features and the dependencies between them are stored within the LSTM units without explicitly needing to know the class of the object.

Our model builds on earlier work before deep learning took off where context was explicitly modeled in the form of semantic context (object co-occurrence), spatial context and scale context ~\cite{rabinovich2007objects,parikh2008appearance}. But unlike these approaches, our model can take into account the semantic context of a set of objects instead of a pair, does not involve separate terms for the classifier probability and context probability which are difficult to fuse and is end-to-end learned. We benchmark the model on the LSUN dataset~\cite{yu2015lsun} which contains 10 million images across 10 categories. The Context-CNN model achieves an accuracy of 89.03\% on the validation set which makes it one of the top performing models on this dataset. We also compare our base network with some standard models and with variations of our model which aim to verify the various assumptions we make about our architecture through control experiments. Additionally, we also analyse the CNN and LSTM features and perform experiments to highlight the context modeling capacity and the discriminative capacity of the model. To summarise, the main contributions of this paper are:
\begin{enumerate}
  \item We propose a new model for scene classification that we refer to as the Context-CNN model which exploits the joint presence of a set of objects in the image to infer the scene category, thus modelling semantic context. We test our model on the LSUN dataset and it produces results comparable to the state-of-art. Additionally, it requires only a small fraction of the LSUN dataset (only ~200k out of the total 10 million images) for the training to converge.
  \item Unlike the previous methods that model context, our model is learned end-to-end. It also models the dependencies between multiple objects at the same time without requiring any object labels.
  \item We perform extensive experiments to demonstrate that the LSTM units used for capturing object-level information are responsible for improving the accuracy. We also analyse various layers of our model empirically and visually to understand the behavior of the network.
\end{enumerate}

The rest of this paper is organized as follows. In the next section we survey related work in the literature. In Section~\ref{context-cnn model} we describe our Context-CNN model with details of the architecture and network training. We present the experimental results in Section~\ref{results} with an analysis of the results through some control experiments. In Section~\ref{visualisation} we analyse our model through visualisations and finally conclude the paper in Section~\ref{conclusion}.

\section{Background \& related work} \label{related work}
Earlier work in image understanding involved numerous low-level image representations like image contours, high contrast points, histogram of oriented gradients \etc ~\cite{tuytelaars2008local} which were carefully hand crafted and proved effective for tasks like object recognition. Other image-level features like GIST~\cite{oliva2001modeling} that capture global image statistics and the spatial information were shown to be effective in recognising scenes. More recently, deep features derived from CNNs have fared extremely well on object recognition tasks, but these CNN architectures have not had the same success with scene classification. The cause of this under-performance is often attributed to the semantic gap between the (local or global) statistical information captured by low-level features and the semantic information required for making scene level decisions. We briefly review some approaches for scene recognition followed by the past work done to bridge this semantic gap in vision tasks finally followed by a review of the previous methods which use the CNN-LSTM model.

\textbf{Scene classification. }
Both global scene descriptors like GIST~\cite{oliva2001modeling} and spatial pyramids~\cite{lazebnik2006beyond} and local, low-level features like SIFT~\cite{lowe2004distinctive} have been used in the past for scene classification. Others part based models like~\cite{pandey2011scene,juneja2013blocks} try to obtain mid-level information from deformable parts. Although image-level features capture the holistic information of the scene, and low and mid level features capture the object information in a scene, the above methods concentrate only on the image statistics and don't attach any clear semantic meaning to a scene or its constituents.

Alternatively, some other methods build on an object-centric view of a scene where a set of objects are the discriminative characteristics of the scene.~\cite{li2010object} uses a scene representation built from pre-trained object detectors. \cite{zhang2014learning} introduces a measure for object-class distance to generalise the idea of an object bank and uses it for classification. In the more recent literature,~\cite{DBLP:journals/corr/WuWWY15} uses discriminative clustering of the deep CNN features of scene patches to form meta objects which pooled together at different scales makes the final scene representation. Various other CNN architectures for scene classification have been proposed in the last few years. MOP-CNN model~\cite{gong2014multi} pools deep CNN features at different scales for smaller image patches and obtains the VLAD descriptor for the entire image. \cite{wang2015training} uses supervision from auxiliary branch classifiers to decide whether or not to increase the depth of the network. A similar technique of using auxiliary supervision layer along with fisher convolution vectors is used for learning the final feature representation in~\cite{guo2016locally}.

\textbf{Context modeling. }The task of utilizing context information for scene understanding has seen a lot of attention. \cite{torralba2003contextual} builds contextual priors based on the position, scale and object categories for learning priming of objects while \cite{torralba2003context} uses an HMM model to incorporate global context.
Co-occurrence of objects, regions and even labels are often used to constrain the learning for various tasks\cite{heitz2008learning, li2012object, mensink2014costa}. \cite{izadinia2014incorporating} uses the scene layout constraint to learn the topology of a scene and perform categorisation. \cite{choi2012context} uses a graphical model to exploit co-occurrence, position, scale and global context which together is used to identify out-of-context objects in a scene. Similar definitions of context are used in \cite{rabinovich2007objects,parikh2008appearance,divvala2009empirical} to model semantic, geometrical and scale context.  \cite{li2011extracting} considers the unlabeled area around a labeled bounding box as contextually relevant and adaptively adjusts the granularity of the regions surrounding a bounding box.
In the more recent vision literature involving deep learning, context has been used in a variety of tasks including pose estimation~\cite{yao2010modeling}, event recognition~\cite{wang2015video}, activity recognition~\cite{rstarcnn2015} and object detection~\cite{gupta2015exploring}. R*CNN~\cite{rstarcnn2015} uses a primary bounding box for recognising the person involved in the activity and a secondary bounding box chosen from multiple contending regions which provides a contextual cue for the identification of the activity. The scores from both boxes are learned together in an end-to-end manner through a R-CNN architecture. Similar to R*CNN, we too use object proposals to obtain object bounding boxes and extract context from them. \cite{wang2015video} builds a deep hierarchical model which exploits context at three levels, namely the semantic level, the prior level and the feature level. This is one of the few models which tries to learn the interaction between the various contextual cues. A multiple instance learning based approach with a VGG network is used in~\cite{gupta2015exploring} to identify regions within an image which may be contextually relevant to the presence of a certain object. These selected regions are then used to reason about the category of the object.

\textbf{CNN-LSTM models. } CNNs have been very successful in learning discriminative features for vision problems and recurrent neural networks have been shown to effectively model the dependencies between its inputs. Many recent architectures use a combination of CNN and LSTM to jointly learn the feature representations and their dependencies. Multi-modal tasks like image captioning~\cite{vinyals2015show,mao2014deep,karpathy2015deep} and visual question answering~\cite{antol2015vqa,ren2015exploring,fukui2016multimodal} use CNN for the image features while the LSTM generates the language for the caption or the answer. Some recent approaches to scene labeling and semantic segmentation use CNN-LSTM architecture~\cite{pinheiro2014recurrent,byeon2015scene,liang2015semantic,liang2016semantic} as CNN-only architectures contain larger receptive fields which do not allow for finer pixel-level label assignment. LSTMs also incorporate dependencies between pixels and improve agreement among their labels. Tasks involving videos also employ LSTM after extracting deep CNN features of individual frames~\cite{wu2015fusing, donahue2015long,yue2015beyond} since the temporal component of videos are suitable inputs for LSTM units. But as some other very recent works show, even in absence of temporal information, CNN-LSTM models can be used effectively to model relationships between image regions or object labels ~\cite{bell2015inside,liang2015recurrent,wang2016cnn}. We borrow from these works to use a CNN-LSTM combination to model context.


\begin{figure}
  \includegraphics[width=\linewidth]{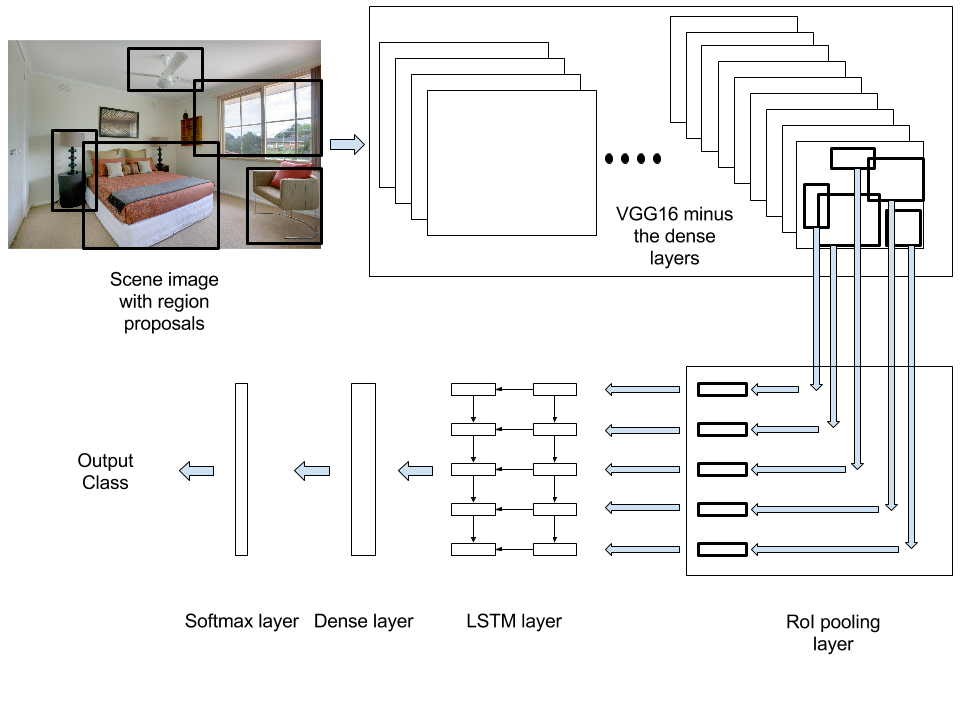}
  \caption{Context-CNN model architecture}
  \label{fig:fig2}
\end{figure}

\section{Context-CNN model} \label{context-cnn model}
The goal of our model is to complement the deep CNN features with high-level semantic context from objects within a scene. The following sections provide the details of the Context-CNN model and its training procedure.

\subsection{Model architecture} \label{model}
Our model (see Figure \ref{fig:fig2}) uses a pre-trained VGG16 network~\cite{Simonyan14c} to extract CNN features but other choices like AlexNet~\cite{krizhevsky2012imagenet} or ResNet~\cite{he2015deep} would work just as well.
The input size of the images are fixed at $512 \times 512$ and the last convolutional layer produces feature maps of size $32 \times 32$. Bounding boxes are extracted using edge boxes~\cite{ZitnickECCV14edgeBoxes} and  the feature maps of these object boxes are passed through an RoI pooling layer~\cite{girshick2015fast} to generate a fixed size vector of size $7 \times 7$ per feature map. These object vectors are passed as input to two subsequent layers of LSTM units containing 1024 and 512 units respectively which model the interaction between these object vectors. The output of all time steps are concatenated to build the final feature vector and fed into the dense layers and then through a softmax layer for prediction.

\textbf{Extraction of deep scene features. }The pre-trained VGG16 model computes convolutional feature vector from which the object RoIs are extracted. The features extracted are learned such that these features exploit object-containing bounding boxes to gain greater discriminative power for scene classification. 
This makes the approach different from training a vanilla CNN on the complete image and then using a separately learned system to model the relationship between the object features.

\textbf{Modeling of object-level context. } Object proposals are obtained from edge boxes using the default parameters as mentioned in their paper ~\cite{ZitnickECCV14edgeBoxes}. The bounding box features are fed into the LSTM in decreasing order of their confidence score with increasing time steps.

Let  $x_t \in \mathbb{R}^{M \times W \times H}$ be output of the RoI pooling layer with $M$ feature maps each of a fixed size $W \times H$.
This will be the input to the LSTM at the $t^{th}$ time step. The definition of our LSTM unit follows that of~\cite{graves2013speech}. Let weights $W$ = \{ ($W^{x}_i, W^{h}_i, b_i$), ($W^{x}_f, W^{h}_f, b_f$), ($W^{x}_o, W^{h}_o, b_o$), ($W^{x}_c, W^{h}_c, b_c$) \} parametrise the four gates of the LSTM unit, namely the input gate, the forget gate, the output gate and the memory cell gate respectively. With~$\sigma$ denoting the sigmoid function and $\tanh$ denoting the hyperbolic tangent function, the mapping applied by the LSTM to its inputs at the four gates are,
\begin{align}
i_t &= \sigma(W^{x}_i x_t + W^{h}_i h_{t-1} + b_i),\\
f_t &= \sigma(W^{x}_f x_t + W^{h}_f h_{t-1} + b_f),\\
o_t &= \sigma(W^{x}_o x_t + W^{h}_o h_{t-1} + b_o) ,~\textrm{and}\\
\tilde{c}_t &= \tanh(W^{x}_c x_t + W^{h}_c h_{t-1} + b_c),
\end{align}
where $i_t, f_t,$ and $o_t$ are the values of input, forget, and output gates at time instance~$t$.
$\tilde{c}_t$ is the intermediate value at memory cell and given the above gate outputs, the updated memory cell at $t$ is given by,
\begin{align}
c_t &= \sigma(f_t \odot c_{t-1} + i_t \odot g_t),
\end{align}
and the output of the $t^{th}$ hidden unit is given by,
\begin{align}
h_t &= o_t \odot \tanh(c_t).
\end{align}
A shortened functional form of the whole unit can be summarised as:
\begin{align}
(c_t, h_t) &= LSTM(x_t, h_{t-1}, c_{t-1}, W)
\end{align}

Thus, with each passing time step, the LSTM reads in an individual object feature vector and updates its memory.
This memory helps the model capture scene context by relating objects occurring in that given scene and distinguishing it from other scenes.
The discriminative capacity of the network improves as the LSTM receives more information with increasing time steps.
LSTMs or Recurrent Neural Networks, in general, are typically used to capture recurrence relationship as is common with sequence data like natural language or speech.
It is interesting to note that LSTMs perform well to capture even the co-occurrence relationship of the various objects appearing in the context of a scene where there is no such recurrence.

\subsection{Training details} \label{training}

The VGG16 CNN is initialised with weights trained on the ImageNet dataset\cite{deng2009imagenet} while the rest of the layers are initialised with the method suggested in~\cite{he2015delving}. Stochastic gradient descent with $0.9$ coefficient for momentum and a batch size of $16$ were used to fine tune the model on images with scene categories as targets. The learning rate is initially set to $\eta=1e^{-3}$ and decayed by a decay factor of $d=1e^{-4}$ according to the following standard policy,
\begin{align}
\eta_{new} = \frac{\eta_{old}}{(1+ i \times d)},
\end{align}
where $i$ denotes the number of iterations that have passed. The learning rate and decay factor are changed to $1e^{-4}$ and $1e^{-3}$ after $10k-15k$ iterations depending on the validation loss. No data augmentation or regularisation is applied and the training is done on a single Nvidia Titan GPU.

The number of object proposals used is fixed to $10$ bounding boxes and we use those with the highest confidence scores. Though the selection of this value can also be done heuristically, we limit our analysis to this value as the primary intention of this work is to highlight the use of CNN-LSTM to model context in scenes. Also, increasing the number of object boxes hinders the analysis of the role of the edge box algorithm in the pipeline. This happens because with large number of bounding boxes, almost the entire feature map is covered and it is difficult to evaluate the significance of edge boxes as an object proposal mechanism (see Section~\ref{role of object proposal} for the corresponding experiment). 

\begin{figure}[t]
  \includegraphics[width=\linewidth]{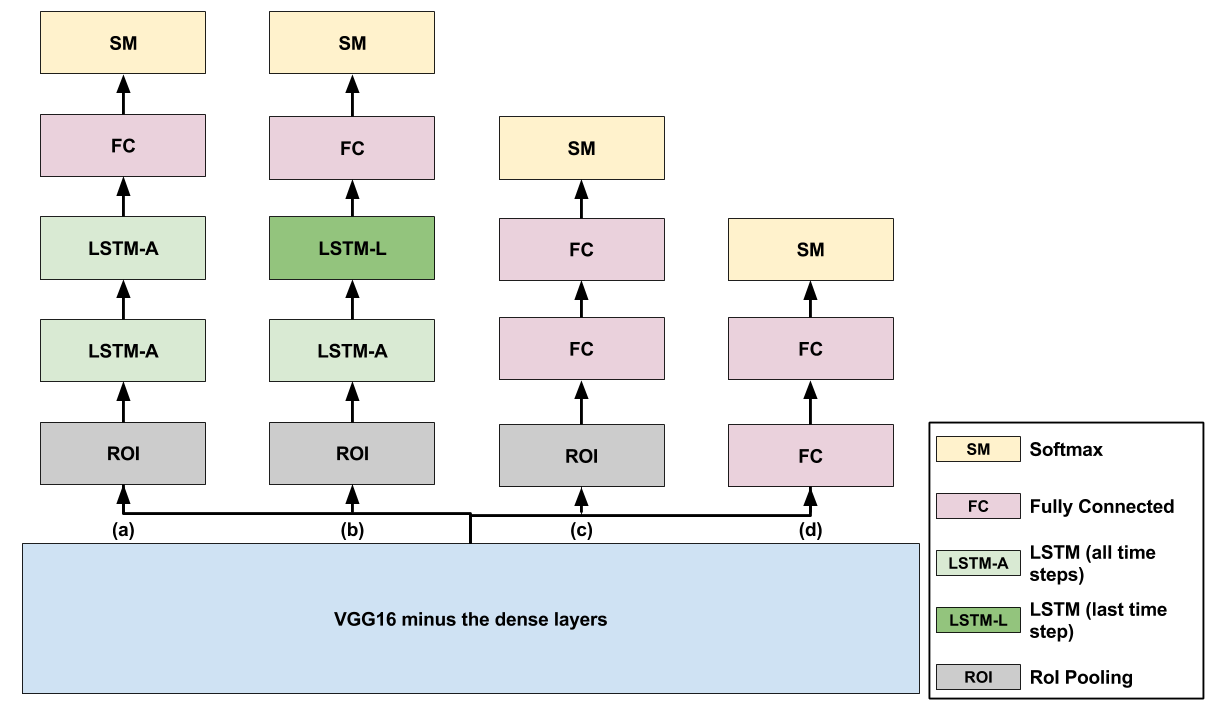}
  \caption{Models comparison: (a) is the base Context-CNN model. (b) shows the the first variation with the output of LSTM coming only from the last time step. (c) shows the second variation with the LSTM units replaced by dense units. (d) is a VGG16 network}
  \label{fig:fig3}
\end{figure}

\section{Experiments \& results} \label{results}
We next describe the experiments and their results  comparing our model's performance on the LSUN dataset with the other state-of-art models.
We also design specific experiments to evaluate and analyze the contribution of object proposals in Section~\ref{role of object proposal} and contribution of LSTMs in the network in Section~\ref{role of lstm}.

\subsection{Results on LSUN dataset} \label{lsun results}
We train and test our model on the LSUN dataset. The best performing variant of our model achieves an accuracy of 89.03\% which is among the best results for this dataset.\footnote{Note that we test the accuracy on the validation set while the official challenge website reports results on the testing set.} Additionally, some models from the leaderboard (see Table~\ref{tab:table1}) use large ensembles, fusing predictions from multiple architectures while we just use a single end-to-end trained model. 

\begin{table}[ht]
\begin{center}
\begin{tabularx}{\linewidth}{|>{\centering\arraybackslash}X || >{\centering\arraybackslash}X|}
\hline
\textbf{Method} & \textbf{Accuracy (\%)}\\
\hline
\hline
SIAT\_MMLAB & 91.61 \\
\hline
Google & 91.20 \\
\hline
SJTU-ReadSense(ensemble) & 90.43 \\
\hline
TEG Rangers(ensemble) & 88.70 \\
\hline
\textbf{Our model} & \textbf{89.03} \\
\hline
\end{tabularx}
\end{center}
\caption{Evaluation on the LSUN dataset}
\label{tab:table1}
\end{table}

To empirically verify the ability of our network to model context, we train variations of our model on the LSUN dataset as control experiments to compare against the base Context-CNN model. The details and results of these experiment are discussed in the following sections.

\begin{figure*}[t]
  \centering  
  \includegraphics[width=0.8\linewidth,height=0.6\linewidth]{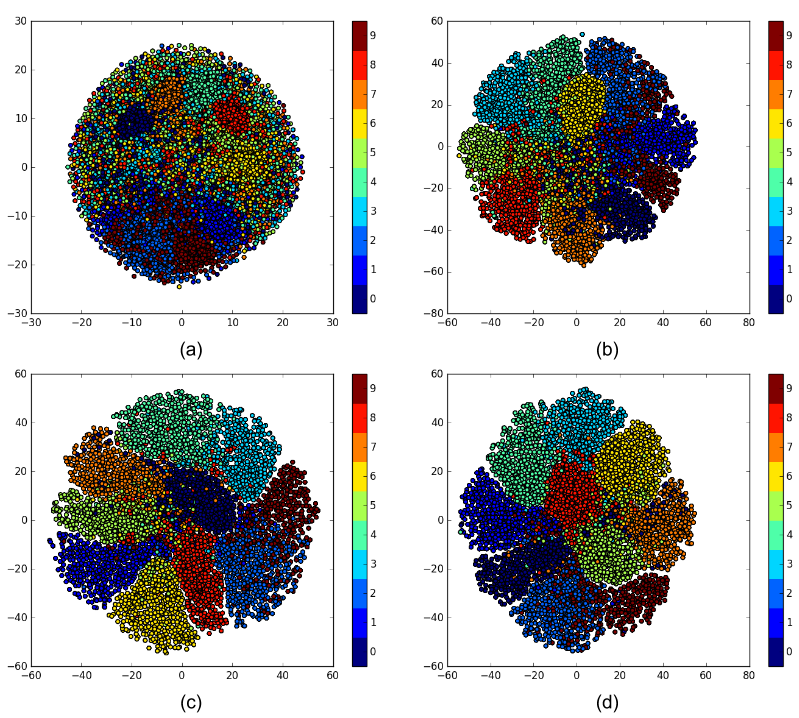}
  \caption{t-SNE visualisation (please view in colour): In (a), each of the data point is a CNN feature vector of a single bounding box obtained from the RoI pooling layer. (b), (c) and (d) show the output feature vector from the $1^{st}$, $5^{th}$ and $10^{th}$ time step of the LSTM respectively. The two axis represent the 2-d plane and the 10 scene classes are denoted by their respective color(see Figure~\ref{fig:fig5} for the names of all classes and their ID). The plot clearly shows how the discriminative ability of the feature vectors of the object bounding boxes change across the CNN and LSTM and also across the various time steps of the LSTM.}
  \label{fig:fig4}
\end{figure*}

\subsection{Significance of LSTM} \label{role of lstm}

We test the base model against three other variations. For the first variation, we only feed the last LSTM time step into the next layer. Since the original model concatenates vectors from all time steps, this variation highlights the importance of the information obtained from the high confidence score objects fed in earlier time steps. It is to be noted that the difference in accuracy between these models is very small as partial effect of the information from the object features of the earlier time steps is reused to compute the later ones.

For the second variation, we replace the LSTM layers by dense layers after the VGG16 model. This setup also includes the RoI pooling layer. 
So in effect, only the object features are modeled, but through dense layers instead of the LSTMs. This model highlights the difference in performance between fully connected units versus LSTM units. The results show that the LSTM units are better at scene discrimination or more precisely, that the LSTM units model dependencies between the deep object features and produce a more discriminative final representation.

For the third variation, we train a simple VGG16 for benchmarking purposes. As expected, it achieves the lowest accuracy out of all the compared variations. We note that even though both VGG16 and Context-CNN share the same convolution layers, they differ in the subsequent layers. So our model outperforms a VGG16 network by 5.6\% with 8 million fewer parameters.

\begin{table}[ht]
\begin{center}
\begin{tabularx}{\linewidth}{|>{\centering\arraybackslash}X || >{\centering\arraybackslash}X|}
\hline
\textbf{Model Variation} & \textbf{Accuracy(\%)}\\
\hline
\hline
Context-CNN base model (Figure 3.a) & 89.03 \\
\hline
Context-CNN with last time step (Figure 3.b) & 87.34 \\
\hline
Context-CNN with LSTM replaced (Figure 3.c) & 85.47 \\
\hline
VGG16 (Figure 3.d) & 83.41 \\
\hline
\end{tabularx}
\end{center}
\caption{Model comparison-role of LSTM}
\label{tab:table2}
\end{table}

\begin{figure*}[t]
  \centering  
  \includegraphics[width=0.8\linewidth,height=0.7\linewidth]{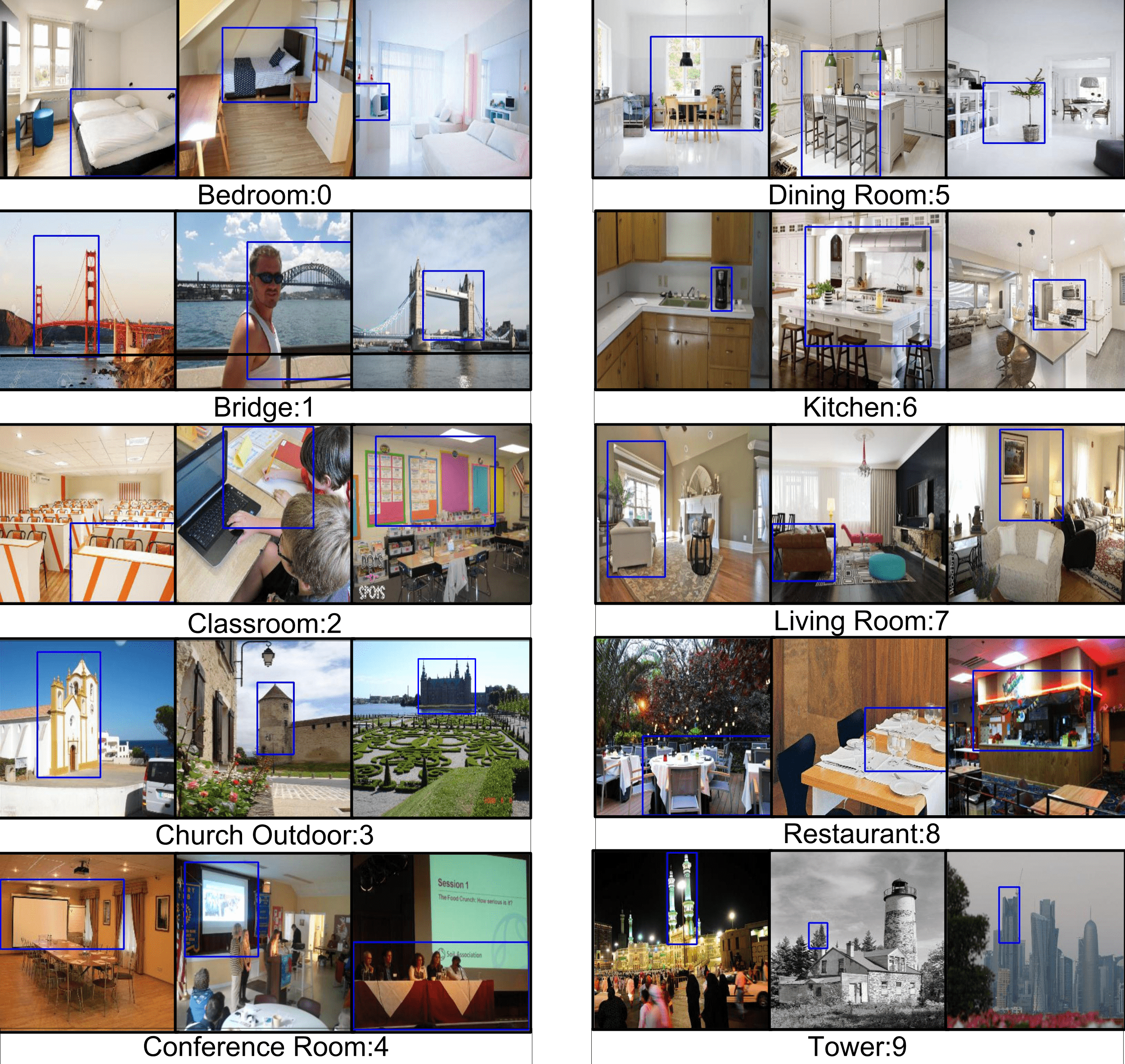}
  \caption{Analysis through obscuration: Systematic blacking out of the object bounding boxes one by one before passing the image through the model and then comparing the obtained softmax distribution to the one obtained with the complete image is used as a measure of significance of the bounding box. The blacked out bounding box which most adversely affects the softmax activation of the correct class are shown for selected images from the LSUN dataset.}
  \label{fig:fig5}
\end{figure*}

\subsection{Significance of object proposals} \label{role of object proposal}

To verify the hypothesis that the proposed network improves scene classification by modeling object-level context, we replace the object proposal method by a mechanism that generates adversarial random boxes.
This mechanism generates random bounding boxes which have similar average size as the original object boxes, but have less than 10\% overlap with any of the original object boxes. An important modification which is made to the Context-CNN model is the increase in size of its feature maps from which the RoIs are pooled. We use transposed convolution layers (also known as fractionally-strided convolution) to upscale the feature map to $256 \times 256$ for this experiment. 
This makes it easier to sample non-overlapping bounding boxes from the feature map and makes the analysis easier. Due to computational limitations, we reduce the number of LSTM and dense units (which together with the upscaling leads to worsening of the results in comparison to the base Context-CNN model), but keep it fixed across this experiment. We report the model with these two changes (the upscaling and the reduction in units) in Table~\ref{tab:table3}. We train two such models, one with bounding box proposals from edge boxes and the other from the adversarial random box generating system.

\begin{table}[ht]
\begin{center}
\begin{tabularx}{\linewidth}{|>{\centering\arraybackslash}X || >{\centering\arraybackslash}X|}
\hline
\textbf{Model Variation} & \textbf{Accuracy(\%)}\\
\hline
\hline
Context-CNN (modified) with edge boxes & 81.56 \\
\hline
Context-CNN (modified) with non-overlapping random boxes & 48.73 \\
\hline
\end{tabularx}
\end{center}
\caption{Model comparison-role of object proposals}
\label{tab:table3}
\end{table}

The results clearly show the difference in performance between the two models. We posit that this large gap in accuracy arises out of the lack of object information in random bounding boxes. 
We note that a similar drop in performance with random boxes is also reported in R*CNN~\cite{rstarcnn2015}, a model which also relies on the presence of objects within bounding box proposals. 
The drop is much more severe in our case since we use adversarial random boxes which have almost no overlap with the original object bounding boxes. 
This experiment verifies the intuition that good object proposals are critical to the modeling of semantic context. These results also imply that the original Context-CNN model is indeed modeling object-level semantic context.

\section{Analysis and visualisation} \label{visualisation}

Experiments from previous section quantitatively measure the contribution of the layers stacked on top of convolution layers. We next give visualizations of the Context-CNN model's feature space and semantically informative image parts that help discriminate between different scenes.

\subsection{Comparison of CNN and LSTM features} \label{cnn vs lstm}

We visualise feature vectors obtained from the CNN and compare it with features obtained from various time steps of the LSTM. We employ t-SNE visualisation~\cite{maaten2008visualizing} to embed the feature vectors obtained from the trained model to a 2-dimensional space. 
Figure~\ref{fig:fig4} shows these embeddings. In Figure~\ref{fig:fig4}(a), each data point represents the CNN feature vector of each RoI-pooled object bounding box. In Figures~\ref{fig:fig4}(b),~\ref{fig:fig4}(c) and~\ref{fig:fig4}(d), each data point is the output from the $1^{st}$, $5^{th}$ and $10^{th}$ LSTM time step respectively. 
The category of the scene from where the object feature vector is taken is used as the category for the feature vector too. Since all features are extracted from a trained network and any particular object feature will encode information about the scene itself, this choice of category for each data point is loosely justified.

The data points in the visualisation can be interpreted as the CNN and LSTM object features. Since each time step takes as input an object feature and modifies its output based on the previous object features, it is expected that with increase in the number of time steps, the capacity of the LSTM features to discriminate among scene classes should also increase. This very intuition is verified visually here.

\begin{figure*}[t]
  \centering
  \includegraphics[width=\linewidth,height=0.4\linewidth]{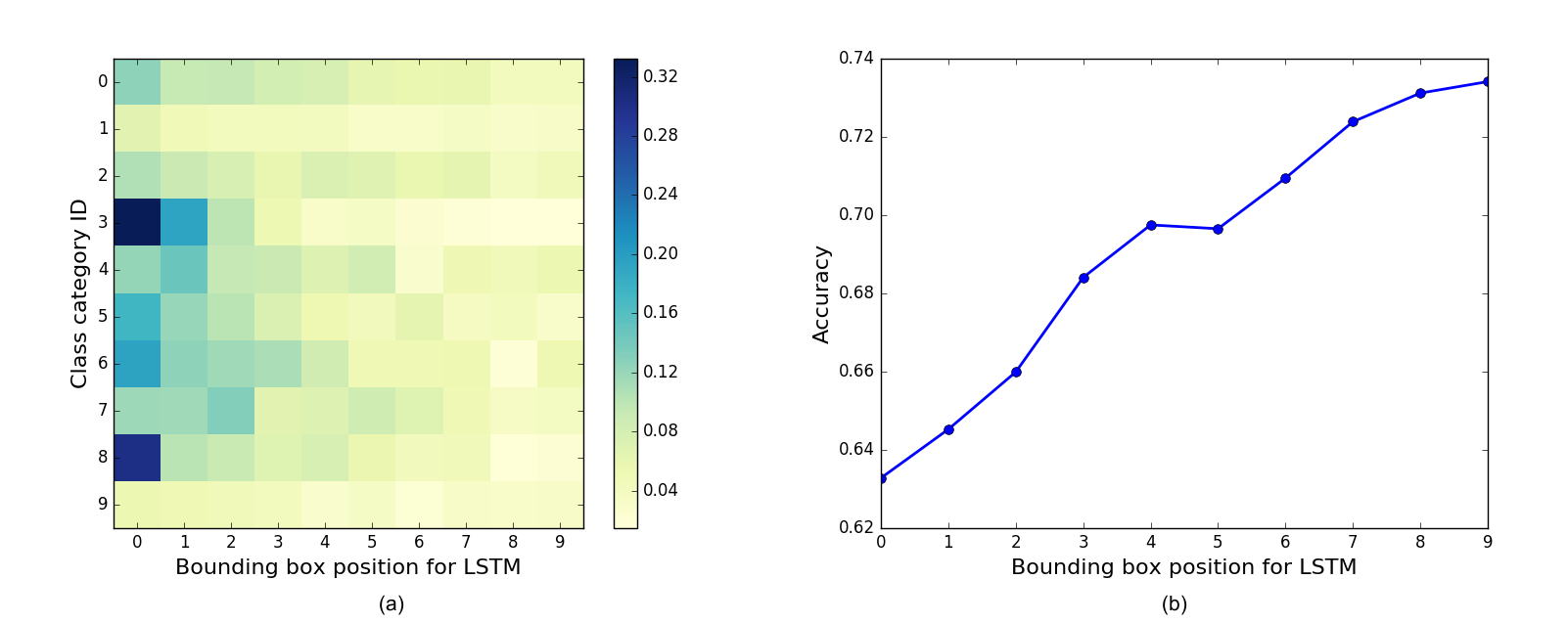}
  \caption{(a)Softmax confidence degradation heatmap: The average drop in softmax scores across the categories and the position of the obscured bounding box with respect to the LSTM time step are plotted as a heatmap(see Figure \ref{fig:fig5} for the names of all classes and their ID). (b)Accuracy of the model as a function of the occluded bounding box position}
  \label{fig:fig6}
\end{figure*}

\subsection{Response to obscuration} \label{obscuration}

Occlusion of various parts of the image was used as a visualisation technique in~\cite{zeiler2014visualizing} to understand which areas of the image contribute how much to the final classification score. 

We take a similar approach to evaluate an object bounding box by defining a measure to quantify its significance for the scene classification task. 
The significance of a bounding box is measured by the reduction in the softmax score of the correct class if the bounding box were obscured and the corresponding object occluded.  The most significant bounding box is the one that leads to maximum reduction in the softmax score.
The best performing Context-CNN model is used for this visualisation. Select representative images of scenes from all categories are shown in Figure~\ref{fig:fig5} and the corresponding observations are as follows:

\begin{itemize}
  \item Bounding boxes which cover a large area, usually, tend to cause the largest reduction in the softmax scores of the correct class when obscured.
  \item Each scene category contains a small set of distinct characteristic objects that help discriminate its images from that of others e.g.\ bed for a bedroom, projector for a conference room, sofa for a living room and so on. Each object from within the characteristic set of a given scene category could, however, vary widely in appearance and pose.
  \item It is sometimes the case with certain scenes that the most significant bounding box is small in size but contains some contextual information which can be exploited in the absence of other discriminatory features. e.g.\ a glass of water for a restaurant scene or a pencil and paper for a classroom scene.
\end{itemize}

We also use obscuration to visualise how the order of feeding bounding boxes into the LSTM could affect the final softmax scores. The drop in scores due to obscuring each bounding box for a given scene category is plotted in Figure~\ref{fig:fig6}.a. As expected, obscuration of edge boxes being fed in the first few time steps reduces the softmax score of the correct class by a greater value than the later time steps. 

This can be attributed to the fact that the initial time steps take as input the bounding box with the highest \textit{'objectness'} score as measured by the edge boxes algorithm. Additionally, some classes like the tower scene and bridge scene show uniform degradation of confidence across LSTM time steps.

Figure~\ref{fig:fig6}.b plots classification accuracy against the time step at which the corresponding occluded bounding box was fed into LSTM layer. As apparent, occlusion of the first bounding box most severely affects performance, dropping the accuracy from 89.03\% (which corresponds to the base Context-CNN model) to 63.3\% (which corresponds to the base model with bounding box at the first time step occluded).

\section{Conclusion} \label{conclusion}
In this paper, we propose a deep model for embedding high-level semantic context of scenes. We evaluate it on the task of scene classification on the LSUN dataset producing results comparable with the best performing methods currently available in the literature. Additional experiments to understand the proposed model point to its effectiveness in modeling relationships between object-level patches of the scene. The results indicate that complex scenes which do not have any globally discriminative features, need to rely on a principled way of joint learning of multi-level representations and objects or image patches are a good way to incorporate these features. The model we propose can also be adopted to other tasks in vision to capture contextual information.

\cleardoublepage
{\small
\bibliographystyle{ieee}

}

\end{document}